# Applied Monocular Reconstruction of Parametric Faces with Domain Engineering

Igor Borovikov, Karine Levonyan, Jon Rein,
Pawel Wrotek and Nitish Victor

Electronic Arts, Redwood City, CA, USA

## Abstract

*Many modern online 3D applications and videogames rely on parametric models of human faces for creating believable avatars. However, manual reproduction of someone's facial likeness with a parametric model is difficult and time-consuming. Machine Learning solution for that task is highly desirable but is also challenging. The paper proposes a novel approach to the so-called Face-to-Parameters problem (F2P for short), aiming to reconstruct a parametric face from a single image. The proposed method utilizes synthetic data, domain decomposition, and domain adaptation for addressing multifaceted challenges in solving the F2P. The open-sourced codebase illustrates our key observations and provides means for quantitative evaluation. The presented approach proves practical in an industrial application; it improves accuracy and allows for more efficient models training. The techniques have the potential to extend to other types of parametric models.*

## Keywords

*Face Reconstruction, Parametric Models, Domain Decomposition, Domain Adaptation.*

## 1. Introduction

Modern virtual environments strive to deliver a life-like representation of human facial likeness under a limited computational budget. Such demand emerges both in the production art tools and user-facing applications. Customizable video games or 3D chat characters generated from user-provided photographs are in high demand (e.g., [13, 19, 23]). The following subsection explains some necessary terminology.

### 1.1. 3DMM vs. Parametric Model

The paper's context requires emphasizing some crucial distinctions in the avatar creation frameworks. Namely, there are two fundamentally distinct methods to creating human faces in Computer Graphics. One approach utilizes a fully Morphable 3D Mesh (3DMM) to adjust individual vertices to produce a desired shape [2]. A radically different approach is parametric (see [2,37]). A parametric model abstracts from the vertices and, in that sense, is more general. It relies on a pre-fixed collection of hand-authored construction elements (sometimes called "blendshapes"). These elements are used across all the characters within an application. Such elements contribute to the target shape with the weights defined by the input parameters, e.g., the distance between eyes, size of the mouth or nose, and alike.

A direct comparison of 3DMM vs. parametric frameworks for ML applications is rarely insightful within an industrial context: the engineering, production, and art style decisions take precedence in designing a concrete application. On the one hand, models can be hand-crafted with highly detailed 3DMMs in the film industry. On the other end, expensive authoring, tight development timeframe, limited memory, or bandwidth get better use with a modest set of predefined assets. Such assets can be preloaded to the customer hardware and allow for compact encoding of the models via a relatively small parameter set in interactive environments.

The paper focuses on *parametric* models suitable for interactive user-facing software and leaves out 3DMM-based systems for a different exploration.

## 1.2. Face-to-Parameters with a Heterogeneous Target Space

In our formulation, the Face-to-Parameters (F2P) problem aims to reproduce the face on a single input image in the best manner possible by optimally selecting two types of model parameters: continuous and discrete. The presence of discrete construction elements makes our target space heterogeneous.

Continuous parameters, as we already mentioned, are the weights of blendshapes. Blendshapes are a standard technique in modeling complex articulated objects, like human bodies and faces (blendshape methods reviewed in [39]). An application or a modeling tool has a fixed set of blendshapes used across all the characters. Given design limitations, the blendshapes aim to represent all anticipated target geometries in the best manner possible. All blendshapes contribute to the resulting geometry simultaneously and are not mutually exclusive. Some examples of a blendshapes effect on the output are the length of the nose, the shape of the nose (curvy, straight, up, down, wide, narrow), mouth location relative to the chin and the nose, the shape of the mouth (plum, thin, curved up or down), how prominent is the chin, and alike.

The discrete elements are fixed too but mutually exclusive within their region of the target geometry. They represent gross deformations of an underlying mesh. Possible examples are a distinct nose shape not reachable with the available blendshapes (say, a broken nose), a particular texture layer (e.g., an artist painted enhanced details for the features too fine to represent with the mesh), or both (e.g., a carnival mask).

In the ML context, fitting the continuous parameters is a regression problem and fitting the discrete elements is a classification problem. In this paper, the F2P problem combines both. In our proprietary application, the dimensionality of the continuous parameters space is ~100, and the discrete elements count exceeds 300. The parameters by design come partitioned into several facial regions (e.g., nose, mouth, and alike), leading to the combinatorial complexity in the order of $10^{11}$. For comparison, the FLAME model [12] has no discrete elements and exposes ~300 continuous parameters, with ~100 devoted to articulating the character. We exclude facial expressions from consideration in this paper. An open-source software Makehuman [15], which we use for reproducible quantitative evaluation of this paper, exposes over 100 continuous parameters and only a handful of discrete ones. Architecturally, Makehuman is the closest counterpart of the proprietary software we used for this work. That dictates our choice to open-source our experimental code using Makehuman rather than other systems like FLAME.

For the brevity of this presentation, we omit color palette elements, hairstyles, facial hair, makeup, tattoo, accessories, etc. - anything that does not directly change the geometry of a character's face. Also, we exclude ears and neck as they are frequently occluded in the imagery.

At the high level, the F2P problem formulation here is similar to that one in [23].

### 1.3. The Approach Outline

For the reasons we explain in the following sections, we cast the heterogeneous F2P problem as a classic supervised learning problem. That requires abundant training data, i.e., facial images with the corresponding parameters. Identifying the source of the training data requires careful navigation around privacy and licensing aspects. Most of the available human face datasets, e.g., a popular CelebA dataset [14], exclude commercial applications. With growing concerns around privacy, many previously accessible facial datasets are no longer available. The most direct way to work around these challenges is to use synthetic data. The generation of synthetic data is often straightforward and can produce practically unlimited amounts of it reasonably quickly with a wealth of the associated metadata.

The synthetic nature of the data creates new opportunities not readily available when working with real-world data. The parameters of the human face model naturally map to different facial regions. In parallel to that grouping, the rendering pipeline can include automated generation of the corresponding semantic segmentation of the synthetic images. Such natural separation leads to domain decomposition: the ML pipeline can become a hierarchical ensemble of models dealing with the general structure of the face and local models that control separate regions. The ensemble allows for smaller models that are less prone to overfitting, can train and execute in parallel utilizing data and model parallelism (e.g., [31] reviews conceptually similar works).

While synthetic data facilitates the decomposition and training of the related ML models, it also presents a challenge due to the inevitable domain gap between synthetic and real-world imagery (a brief introduction to domain adaptation in [40]). The imagery such as selfies may be only one of the possible input types to the F2P models. Examples of other possible target domains include sketches, fine art portraits, faces from comics, anime, and more. Previously unseen domains may degrade a naive F2P ML model's accuracy or render it unusable. That makes the domain gap issue broad and even more critical, suggesting that domain adaptation must be an integral part of the system we build. At the same time, the domain adaptation functionality must be modular and easily replaceable to accommodate different future applications.

We utilize style transfer for domain adaptation (see the seminal paper [4]). The direction of the style transfer is from the target imagery (e.g., selfies) to the synthetic images, which could have a distinct (fixed) art style. It is "inverse" in that sense. We train GANs specific to each of the multiple target domains and use them as an adapter. Training such GANs is independent of the decomposition-based ensemble training. We find that the decomposition and domain adaptation enhance each other and lead to an ensemble that produces better results in solving the F2P problem than a direct approach based on a single monolithic model.

### 1.4. Contribution

The paper emphasizes practical aspects of converting facial images into the parameters of a parametric model of a human face and proposes a novel, efficient solution. While the individual building blocks of our approach are not new (hierarchical decomposition, domain adaptation, models ensemble), their proposed combination is not found in the literature.

Concretely, we cast the F2P problem as a supervised learning problem on synthetic data, introduce a model ensemble to take advantage of the hierarchical nature of the domain, and train separate dedicated models for domain adaptation. The proposed architecture is different from the previous works (e.g., [23]) and offers several practical advantages in the industrial environment. The open-sourced code illustrates the claim that the proposed ensemble performs better than a monolithic model thanks to leveraging the structure of the application domain. We advocate the modularity of the proposed system and offer a quantitative evaluation of claims.

In the rest of the paper, we review some of the previous works and then focus on the main objective: building an efficient ML system solving the outlined F2P problem, which infers the target face parameters from the input images. The system we develop is implemented as proprietary industrial software. However, we address reproducibility and quantitative evaluation of the techniques with Makehuman (free open-source software) by open-sourcing our experimental code as well [38].

## 2. PREVIOUS WORKS

Facial likeness reconstruction in computer vision is a vast active area of research. For a relatively recent comprehensive review of the field, we refer to [35]. Here, we highlight only several of the relevant publications. A well-established body of work aims at generating sculptable (and texturable) mesh directly from monocular input: e.g., [18,24,28,19,29,11,22] or video [26,27]. Parametric models may also utilize morphable meshes with a fixed topology where the parametric space explicitly encodes deltas for vertices subsets. Their reconstruction from 2D imagery achieves a very high accuracy [20] but conceptually is similar to 3DMM reconstruction. Such models differ from more specialized ones that use higher-level construction elements, e.g., blendshapes and bone-driven morphs, which are usually specific to a particular graphics engine. The models relying on high-level construction elements are somewhat less common in the literature even though they date back to the early days of Computer Graphics and Computer Vision, e.g., [16]. They provide a critical advantage in data compression, which motivated one of the early works [3]. Compact data representation remains in demand for interactive graphics applications as they strive for computational efficiency. Also, as the result of compact representation, parametric models allow for easier manual authoring compared to freely deformable meshes.

Parametric models heavily depend on the underlying character modeling system's proprietary nature. That creates an obstacle to transferring the parametric techniques across applications and renders a comparison to the 3DMM-based methods irrelevant. The additional difficulty comes from discrete elements of facial reconstruction. In our work, we must handle both continuous and discrete elements. The paper [23] uses a custom-trained differentiable renderer (DR), a powerful tool in the continuous domain. However, a DR could be more problematic for fitting discrete models and is still too heavy for mobile devices.

The paper [23] works around the domain gap between real and synthetic faces by introducing discriminative loss – a perceptual distance computed with embeddings of a facial recognition CNN. The discriminative part becomes an integral part of a larger model. For our purposes of building an ensemble with pluggable components, we explored several other more general GAN-based techniques. Our brief overview starts with [32] discussing various ways to perform domain adaptation. The Neural Style Transfer [4] is the pioneering work utilizing both content and style losses. The approach uses only a single image to represent the style. That appears to be rather limiting for our application. An early generative model applied for image-to-image translation between domains is pix2pix [7], and it uses conditional GANs. The conditional GAN introduces two extra losses [8] that address radical differences for the domains in question. However, it requires paired image-to-image translation. CycleGAN [34] works for unpaired data. It translates the source image into the target without paired examples by introducing the (computationally intensive) consistency loss. More recent advancement on style transfer is well-established StyleGAN2 [30] and its later modification StyleGAN3 [36]. The StyleGAN2 architecture has multiple advantages: the scale-based hierarchy for the generator and the discriminator, pretrained models, and better stability than many other style transfer approaches. Its hierarchical nature helps to control the enhancement of the appearance of stylized images via the so-called layer-swapping. It allows for balancing features at different levels of details using blending weights, making it an attractive candidate for our exploration.

In conclusion of this section, we note that [23] has similar objectives to ours and some superficial similarities of techniques. However, the constraints and the methodology proposed here are quite different. Our target platform excludes a differentiable renderer from the potential toolset. We exclude pixel loss from consideration since we allow various projections and lighting conditions for the input images. Finally, the proprietary codebase of [23] makes reproduction and a direct comparison with its results difficult. As such, the cited work serves more as a valuable inspiration rather than a technical reference.

## 3. F2P AS A SUPERVISED ML PROBLEM

In this paper, we treat F2P as a classic supervised ML problem. The training and validation data are synthetic, generated by the target software. That eliminates any licensing or privacy concerns. The open-source Makehuman [15] provides similar functionality to our proprietary software, and we use it for the quantitative evaluation of this study with the codebase [38].

A simple way to generate the training data is to do it offline. An instrumented client application accepts a "recipe" (a complete set of parameters, including those not considered as target variables) for constructing a character as an input. Next, it renders and saves several views of the generated character. The views may vary by camera, lighting, pose, facial expression, and gaze direction. Here, we only limit the target space to the facial parameters and exclude the view and pose parameters. The normalized continuous parameters (blendshape weights) map to the floating-point target vector. The discrete components use one-hot encoding occupying target vector slices of the size corresponding to the number of the options.

A direct approach to training a model mapping an image to the heterogeneous target is to train a CNN with multi-part loss function $L$:

$$L = \sum_{i=1}^{N} v_i R_i + \sum_{i=1}^{N} w_i C_i \qquad (1)$$

Here, $N$ is the number of facial regions (seven in our main case study and we keep only three for Makehuman experimentation). $R_i$ is either mean L2 or L1 loss for modifiers, $C_i$ is the cross-entropy loss for the discrete elements, and $v_i$, and $w_i$ weights that can be adaptive [5]. We use transfer learning with *inception3* [25] and *squeezenet* [6,33] from Pytorch [17] models zoo as the starting point. Transfer learning provides a reasonable accuracy relatively early in the training process, but complete training may take longer with ~10k input images randomized across all target dimensions. The relatively long training times in the direct approach are a disadvantage slowing iterations on the models for the evolving customer product. Also, a monolithic model makes it hard to address inaccuracies in concrete facial regions. Finally, the amount of training data may be insufficient for reliable generalization due to the high dimensionality of the target space, potentially leading to various biases and overfitting. These considerations motivated our decomposition approach.

## 4. ENSEMBLE ENGINEERING VIA DOMAIN DECOMPOSITION

The natural subdivision of a human face into regions like the nose, mouth, and alike leads to the target space's corresponding decomposition. Such decomposition is present in many parametric-based modeling applications, including Makehuman. Equation (1), grouped by region, gives loss functions for local models with the following caveats.

The first caveat is overcompleteness, i.e., a parametric model may generate the same visual appearance with different parameter values. One obvious source of such over-completeness is scale. Fixing a particular scale variable in the training data normalizes it and eliminates over-completeness in our studies. Next, a group loss corresponding to a specific facial region may include local (e.g., angle) and global features (e.g., placement of the feature relative to the other facial features). Manual engineering of the group loss to account for such subdivision could be error-prone and subject to frequent revisions as the client software evolves. We address local-vs-global ambiguity by introducing weights into the ensemble as follows. Global features learned within local groups would result in predictions equal to an average value over the dataset. When learned as part of the aggregate complete model, they will result in a much better prediction. Their learned combination with the introduced weights will automatically reflect their roles. One way to formulate training for the ensemble weights is to frame it as an optimization problem

$$E(w) = |\sum_{i=1}^{n_k} w_i^{(k)} M_i^{(k)}(D) - T(D)|^2 \to \min_w \qquad (2)$$

Here, $E(w)$ is the cumulative L2 error computed for weights $w^{(k)}_i$ corresponding to the group $k$ and target variable $i$. By running models $M^{(k)}$ on the training dataset $D$, we obtain predictions for each group k. Linear combination of the predictions with weights $w^{(k)}_i$ gives ensemble prediction. We compare it with the known target vector $T(D)$ and compute mismatch on the entire dataset as a function of $w$. The weights vectors $w^{(k)}_i$ dimensionality equals the dimensionality of the target parameters space. Also, we normalize the weights so that they sum to 1 for each target variable. That gives complementary weights $w_i$ and $1-w_i$ for the coordinates shared by local and global features. Solving (2) for $w$ is a straightforward coordinate-wise task. Not restricting weights to positive values allows to utilize consistent bias in either local or global models and can further improve the accuracy of the ensemble. Figure 1 summarizes our approach to constructing, training, and using the decomposition-based ensemble for inference.

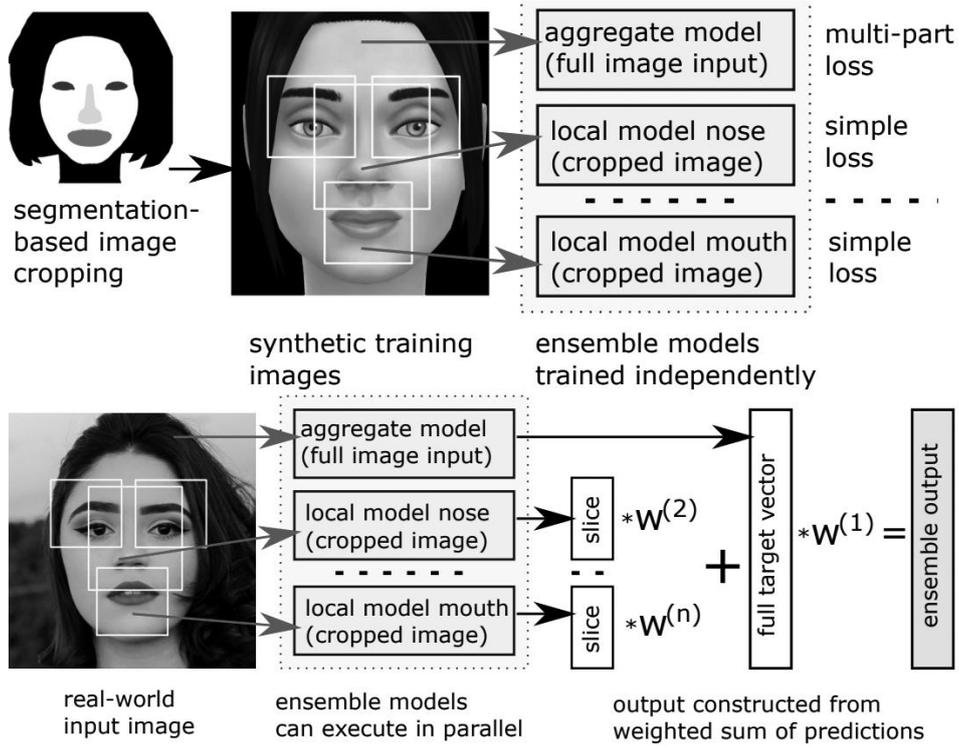

Figure 1. Training (top) and inference (bottom) with the decomposition-based ensemble.
(We defer domain adaptation part until the upcoming section, Figure 5.)

A complete target space in our study has dimensionality close to 400. Its combinatorial part has a complexity of ~$10^{11}$. Decomposition into subspaces radically reduces the dimensionality of the target spaces for each model we need to train. In our application, the largest modifiers subspace has dimensionality ~20. The largest number of discrete options for discrete elements for a region is under 100. Since the subspaces for discrete parts represent mutually exclusive choices, the combinatorial complexity reduces by many orders of magnitude. Combined with lower input resolution for local features, these factors allow for more efficient training with a smaller amount of training data necessary for sufficient sampling coverage of the target space. Since training models for the decomposed local features are less demanding than for an aggregate model, we may use smaller CNNs. In our experiments, we use the Pytorch model zoo's *squeezenet* for local features inference. Its smaller input resolution, 224x224, is suitable for cropped input image parts. Also, *squeezenet* is a lightweight model at only ~3Mb vs. *inception3*, which is close to 100Mb and takes 299x299 inputs. Our main study has 13 models, two per region (regression and classification grouped separately), with one region lacking discrete elements.

## 5. ENSEMBLE EVALUATION WITH AN IN-HOUSE SOFTWARE

In our first round of experimentation, the proprietary software has a stylized, cartoonish rendering, defined by the art direction. Using automated objective evaluation (e.g., cosine distance in the latent space of FaceNet [21]) while disregarding the artistic style is problematic due to the domain gap. All our training and evaluation datasets are synthetic, rendered in the same artistic style, while we expect that input images would come from photographs. To workaround, we use a panel-of-experts method. Nine experts evaluate the reconstruction quality of a hand-picked set of unannotated 20 facial images representing various ages, ethnicities, lighting, image quality, poses, and facial expressions. During the evaluation, the respondents, besides other things, have to rate each of the seven major facial regions of the reconstructed characters with binary good/bad evaluation (could be less informative than the Likert scale but makes the questionary easier and faster to complete).

Table 1 summarizes the nine experts' responses in both rounds of the assessment: one for the naive aggregate model and the other for the models' ensemble. The panel includes both professional artists and technical staff members. The table highlights improvements in the selected quality metrics obtained by introducing the model ensemble and clearly shows the advantages gained from the decomposition approach.

Table 1. Comparison of ensemble vs. aggregate models by a panel of experts. Models' decomposition and ensembling reduce reconstruction defects by 30-50%. The worst possible defects score is 180 for nine experts and 20 images. The FaceNet cosine distance to the input image also reduces on most images by ~ 10%.

| **Reconstruction defects by region** | Cheeks | Chin | Eyes | Forehead | Jaw | Mouth | Nose |
|---|---|---|---|---|---|---|---|
| Naive aggregate model | 19 | 18 | 69 | 3 | 29 | 37 | 34 |
| Ensemble with models' decomposition | 6 | 11 | 34 | 1 | 16 | 24 | 17 |

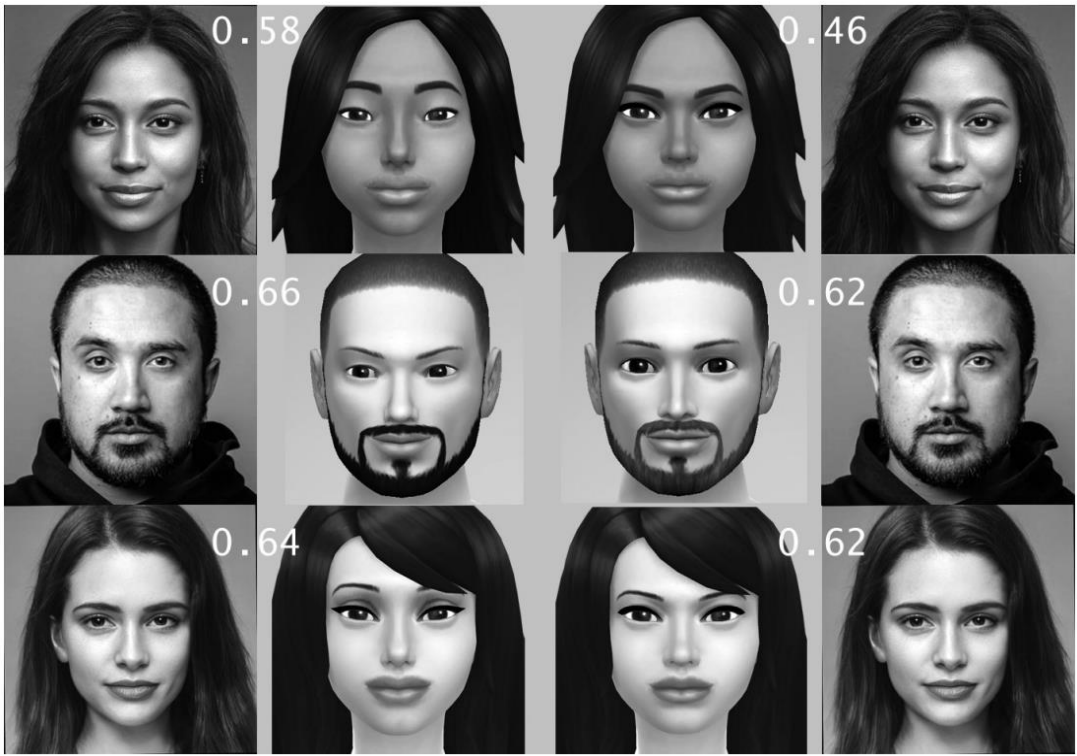

Figure 2. Naive aggregate model reconstruction (second on the left column) compared to the decomposition with models ensemble (second on the right). The numbers indicate FaceNet [21] cosine distance between the corresponding pairs. The imagery does not intend to represent any current or future commercial product. Photography attribution (side columns) [1].

## 6. DOMAIN GAP AND DOMAIN ADAPTATION

Naturally, we cannot ensure that synthetic training imagery covers the entire domain of the expected inputs or is its representative sub-domain. Also, in applications, we may encounter new domains not anticipated during training. The domain gap between synthetic and target imagery makes the models underperform.

Two major factors are leading to the domain gap. One is the limitations of the parametric model itself, which may not be powerful enough to generate a sufficient variety of faces. The second factor is the artistic stylization produced by the client application. The stylization lends itself to various domain adaptation techniques. Using style adaptation (e.g., see [4]) as an intermediate step between the input and the rest of the inference pipeline should reduce accuracy loss. Namely, we propose using an ``inverse'' style adaptation from the input to stylized synthetic imagery. We can train such an adapter for various inputs besides real-world images, e.g., fine art, anime, sketches.

After assessing various style transfer options (see Previous Works section), we use StyleGAN2 [30] for the academic part of our study (it has no commercial license) to train and evaluate the proposed style adapter in application to our F2P inference pipeline. The synthetic training domain in our experiments comes from Makehuman [15], picked for its architectural similarity with our proprietary software. The target domain is real-world imagery. For inference, we apply StyleGAN2 to the input image to make it look like it comes from Makehuman, then feed it into the pipeline trained with Makehuman synthetic dataset. To train the adapter, we start with a StyleGAN2 pretrained on the FFHQ dataset (created for [9]) and fine-tune it on 4000 Makehuman-generated images. We normalize the images by registering (resizing and aligning) them using dlib landmarks [10] to match the images' alignment in the FFHQ dataset. After computing the style weights, the inference continues as follows. We start with a normalized real image. Next, we compute the latent projection vector for the given image through the StyleGAN2 mapping network and then apply the blended style weights from the selected resolution to map the real image to the stylized image. Figure 3 illustrates the results of that process. Finally, we feed the stylized image to the inference F2P pipeline, which becomes an ensemble including the domain adaptation step.

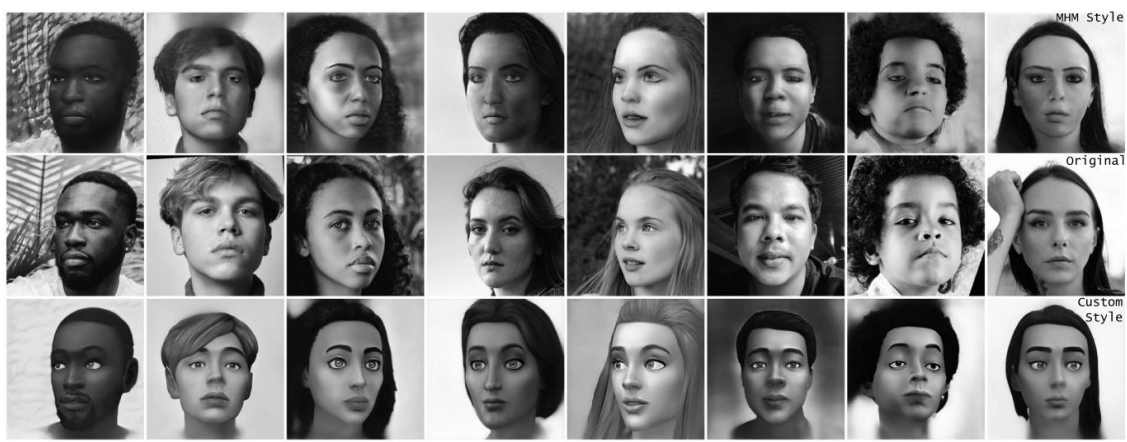

Figure 3. Stylized images from unsplash.com to the proprietary and Makehuman styles. Top row: Makehuman style, middle: original photos, Bottom row: proprietary style.

## 7. QUANTITATIVE METRICS AND ABLATION STUDY

We conduct a series of experiments with Makehuman [15] software, reproducing the setup of our proprietary-based experimentation to support our findings using open-sourced, reproducible, quantifiable methods. For simplicity, we use a trivial baseline for the collected metrics: a mean over the evaluation dataset for the target variables. In our case, it corresponds to zero for all target variables describing an average face shape. Here, we focus on the regression part of the problem since Makehuman does not offer many discrete elements for the classification part. Besides, the decomposition advantage for classifiers follows directly from the combinatorial considerations.

Note that our goal for this concrete open-source study is not to train the best CNNs possible. That requires more effort and experimentation. Instead, we fix the training setup (meta-parameters, adaptive multi-part loss tuning, and the training schedule) to a reasonable common one and compare results in terms of accuracy between the combinations of the binary factors we describe after stating the following two claims to verify:

- **A decomposition-based ensemble** improves inference accuracy over the monolithic model.
- **Style adaptation via inverse style transfer** as a preprocessing step improves the accuracy of a model or models trained with synthetic training data.

For the decomposition-based ensemble claim, the goal is to obtain metrics characterizing the accuracy of the models trained either as a monolithic or decomposition-based ensemble. We evaluate the models utilizing the pretrained *squeezenet*, which we update in feature extraction and fine-tuning transfer learning modes. The motivation for considering both comes shortly. Overall, these three binary factors influence the results:

1. **Target space partitioning:** a complete target vector or limited to a particular feature group (e.g., nose parameters only). Such partitioning corresponds to the grouping terms of the target variables by features in multi-part loss (1).
2. **Input image:** either a complete frame or the corresponding to a feature group "semantic" crop. We resize the cropped input image to match the CNN input.
3. **Transfer learning mode:** feature extraction or fine-tuning. In our tests, the fine-tuning phase starts from the CNN obtained by the features extraction step to speed up training.

Figure 4 illustrates the setup for binary factors we test, and Table 2 summarizes the results.

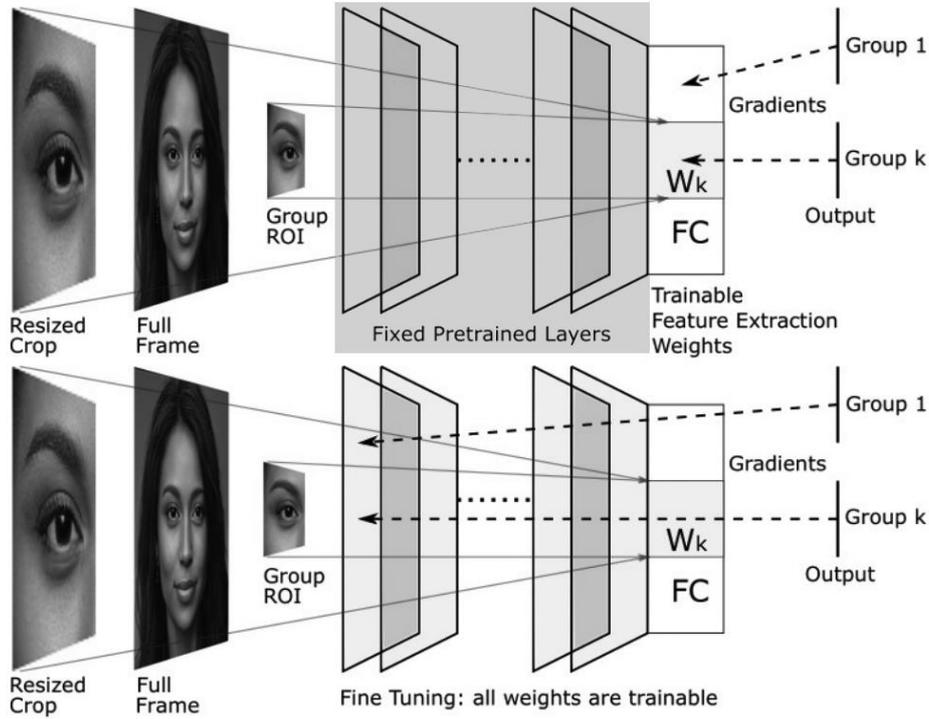

Figure 4. Quantitative evaluation setups and ablation study. We compare pairs: full-frame vs. crop input and complete multi-part loss vs. local group loss. Top: transfer learning with feature extraction. Bottom: transfer learning with fine-tuning.

We start with the target space partitioning. One of the factors that may influence the models' training and the resulting accuracy is the difference in scale of the gradients in the multi-part loss (1). If such a difference is significant, it may take longer to achieve the required accuracy across all target variables. We train the first set of models using feature extraction transfer learning to isolate that factor. In that form, the CNN's hidden layers' weights remain fixed with only the top fully-connected (FC) layer trained to fit the dataset. The input in this experiment is a full-frame image for all models.

When training stops with a predefined learning rate schedule by reaching a plateau for the evaluation dataset, we expect similar results between the local and monolithic models compared to the local target features. Running training sufficiently long with a shared learning rate scheduler should reduce the different gradient scales' effect (regardless of the overfitting concerns). The top part of Figure 4 illustrates the proposed setup. The hidden layers inside the greyed box remain fixed. The grey FC weights $W_k$ for the target group $k$ train (nearly) independently in monolithic and local features variants. The main expected difference is the loss improvement rate but similar resulting accuracy. Hence, the decomposition-based ensemble should not provide a notable advantage over the monolithic model with feature extraction training with a full-frame input. Table 2 shows only a tiny improvement from complete to partial loss function when using the full-frame as an input, supporting the intuition.

Table 2. Local feature group models demonstrate the advantages of the hierarchical decomposition of the F2P problem. The inaccuracy (mean L1- loss) generally decreases left-right top-down within each group. The inaccuracy shown is relative to the baseline. Limiting the number of terms in multi-part loss and cropping the feature improves the resulting model by order of magnitude for some features. The resulting models' ensemble is far superior to the aggregate model trained on the complete input image while sharing with the aggregate one a similar training setup.

| Features Group | Loss | Input | Inaccuracy vs. baseline (smaller is better) | |
| --- | --- | --- | --- | --- |
| | | | Feature Extraction | Fine Tuning |
| Nose | Complete | Full frame | 0.0001 | -0.0039 |
| | Local | Full frame | -0.0005 | 0.0007 |
| | | Cropped | -0.0059 | **-0.0743** |
| Mouth | Complete | Full frame | -0.0001 | -0.0135 |
| | Local | Full frame | 0.0005 | 0.0000 |
| | | Cropped | -0.0080 | **-0.0744** |
| Eyes | Complete | Full frame | -0.0003 | -0.0235 |
| | Local | Full frame | -0.0001 | 0.0010 |
| | | Cropped | -0.0061 | **-0.0530** |

The other factor influencing the accuracy of the models is choosing the input image used to train the local feature CNN. We expect the results to improve by moving from a complete full frame to crops specific to the particular feature groups. That should work even in the feature-extraction transfer learning case. The Feature Extraction column in Table 2 confirms such an expectation. E.g., for the Nose features group, compare numbers in the feature extraction column between Full-Frame and Crop rows.

The progression from complete to local loss while using full-frame input does not improve accuracy much compared to the local loss and cropped images. Moving from feature extraction to fine-tuning, we adjust all weights in the CNN; see the bottom row of Figure 4. That makes learning the target features less constrained and provides a significant boost to the models' accuracy trained with local loss function corresponding to the feature image crop.

Moving from the full-frame to the cropped image input provides the most improvement for fine-tuning. The bold numbers in Table 2 correspond to the proposed local models used for constructing the decomposition-based ensemble. They show the best accuracy across the evaluated combinations of the binary factors.

Table 3. Fitting weights for the features individually reduces the prediction error compared to the simple ensemble with constant weights across all the dimensions. We set the baseline for this table per training type as the corresponding accuracy (validation loss) computed with constant weights (i.e., 0.0, 0.5, and 1.0). Smaller is better.

| Constant weights | Feature Extraction | | Fine Tuning | |
| --- | --- | --- | --- | --- |
| | Full Frame | Crop | Full Frame | Crop |
| 0.0 (aggregate model only) | -0.08 | -0.09 | -0.11 | -0.12 |
| 0.5 (equal mix of aggregate and local models) | -0.08 | -0.11 | -0.23 | -0.27 |
| 1.0 (local models only) | -0.08 | -0.19 | -0.44 | -0.52 |

Finally, Table 3 shows that an ensemble with learned weights for its individual parameters outperforms its components or a weighted sum with fixed equal weights for the predictions.

We conclude this section with a discussion of domain adaptation. The inverse style transfer preprocessing step with StyleGAN2 trained on Makehuman images improves the ensemble's accuracy. We could not utilize validation loss for that evaluation. Instead, we use cosine distance on FaceNet embeddings [20]. The absolute distances shown in Table 4 may look substandard than the commonly accepted threshold of 0.51 for person identification. However, their values are meaningful only in relative terms and illustrate that the domain adaptation moves the input distribution in the right direction, beneficial for the model ensemble. The bottom row distances for full-frame and crop cases are smaller for the style-adjusted images than for the original ones.

Table 4. FaceNet cosine distance to the original input image shows the advantages of the style transfer as the domain adaptation step. Here we use a subset of CelebA [14].

| FaceNet cosine distance | Input | Feature Extraction | | Fine Tuning | |
|---|---|---|---|---|---|
| | | Full Frame | Crop | Full Frame | Crop |
| Original | [0.0] | $0.889_{\pm 0.13}$ | $0.893_{\pm 0.12}$ | $0.917_{\pm 0.12}$ | $0.875_{\pm 0.12}$ |
| Stylized | $0.495_{\pm 0.11}$ | $0.889_{\pm 0.10}$ | $0.895_{\pm 0.11}$ | $0.892_{\pm 0.11}$ | $0.868_{\pm 0.12}$ |

The presented experiments support our claims that in addition to being more manageable, easier to train, the ensemble-based approach takes advantage of the underlying properties of the models we use. It helps to train a better accuracy inference pipeline more practically.

## 8. DISCUSSION AND FUTURE WORK

For completeness, we mention here the limitations of the proposed technique. Reconstruction of a human face in the F2P problem is a multi-faceted task. The paper covers only a single subject - decomposition of "geometric" features that include continuous and discrete elements. In our experience, the classification of glasses, earrings, and some other localized features also benefit from the segmentation and decomposition approach. However, many "spread" features (e.g., hairstyle) do not easily lend themselves to the proposed method. Also, an assumption of the target features' independence is an oversimplification. The human faces exhibit strong correlations between age, gender, and ethnicity features. Including such correlations in the proposed ensemble may further improve the accuracy. One way of doing it is using a Bayesian framework over local models. It can replace a simple weighted sum with a belief propagation network and integrate otherwise ignored correlations between local features. It appears to be a valid subject to explore next. Other parametric models besides human faces may benefit from the proposed methodology.

## 9. CONCLUSION

The paper proposes a novel combination of well-established domain manipulation techniques summarized on Figure 5. Despite its conceptual simplicity, the domain decomposition combined with domain adaptation provides several measurable benefits in the F2P problem that are particularly valuable in the applications to both external user-facing software and the in-house art production pipelines. It facilitates training of the models, offers better control over their accuracy, convenient maintenance vital for industrial applications, smaller memory footprint during inference, and flexibility across input domains. The proposed approach is not fundamentally limited to parametric faces and may work for similar problems in other computer vision applications.

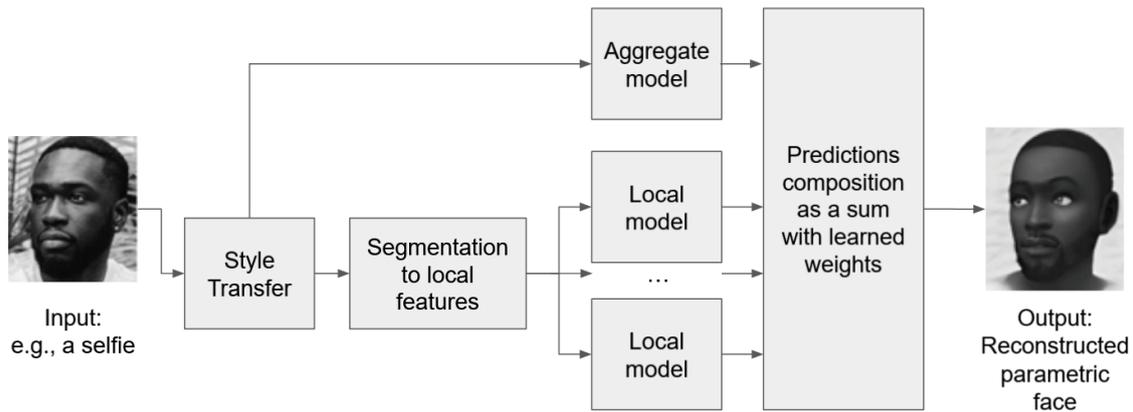

Figure 5. The diagram summarizes the proposed inference pipeline based on domain adaptation and decomposition ensemble. Style transfer, like the local models, is a pluggable module with the corresponding model easily re-trainable as the requirements change.

**AUTHORS**

**Igor Borovikov** received his Ph.D. in math at Moscow Institute for Physics and Technology in 1989. He is a Senior AI Scientist at Electronic Arts.

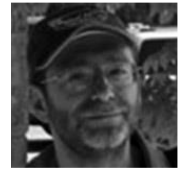

**Karine Levonyan** received her Ph.D. at Stanford University in 2019. She is an AI Scientist III at Electronic Arts.

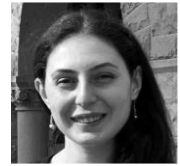

**Jon Rein** graduated from Art Institute (CDIS) in 2004. Currently, he is a Senior Software Engineer at Electronic Arts

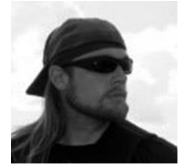

**Pawel Wrotek** graduated from Brown University in 2006 and received M.S. in Computer Science. He is currently a Senior Software Engineer at Electronic Arts

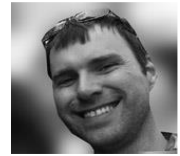

**Nitish Victor** received his M.S. from Rochester Institute of Technology in Game Design and Development in 2019. He is currently a Software Engineer II at Electronic Arts.

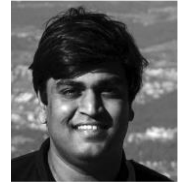